\documentclass[sigconf]{acmart}

\AtBeginDocument{%
  }

\setcopyright{acmlicensed}
\copyrightyear{2018}
\acmYear{2018}
\acmDOI{XXXXXXX.XXXXXXX}

\acmConference[Conference acronym 'XX]{Make sure to enter the correct
  conference title from your rights confirmation emai}{June 03--05,
  2018}{Woodstock, NY}
\acmISBN{978-1-4503-XXXX-X/18/06}

\copyrightyear{2024}
\acmYear{2024}
\setcopyright{acmlicensed}\acmConference[MM '24]{Proceedings of the 32nd ACM International Conference on Multimedia}{October 28-November 1, 2024}{Melbourne, VIC, Australia}
\acmBooktitle{Proceedings of the 32nd ACM International Conference on Multimedia (MM '24), October 28-November 1, 2024, Melbourne, VIC, Australia}
\acmDOI{10.1145/3664647.3681238}
\acmISBN{979-8-4007-0686-8/24/10}

\usepackage[capitalize]{cleveref}
\crefname{section}{Sec.}{Secs.}
\Crefname{section}{Section}{Sections}
\Crefname{table}{Table}{Tables}
\crefname{table}{Tab.}{Tabs.}

\usepackage{multirow}
\usepackage{multicol}
\usepackage{hhline}
\usepackage{tabularx}
\begin{document}

\title{FD2Talk: Towards Generalized Talking Head Generation with Facial Decoupled Diffusion Model}

\author{Ziyu Yao}
\email{yaozy@stu.pku.edu.cn}
\orcid{0000-0003-1310-0169}
\affiliation{
  \institution{Peking University}
  \city{Beijing}
  \country{China}
}

\author{Xuxin Cheng}
\email{chengxx@stu.pku.edu.cn}
\orcid{0009-0002-6244-2931}
\affiliation{
  \institution{Peking University}
  \city{Beijing}
  \country{China}
}

\author{Zhiqi Huang}
\authornote{Corresponding author}
\email{zhiqihuang@pku.edu.cn}
\orcid{0000-0003-1126-1217}
\affiliation{
  \institution{Peking University}
  \city{Beijing}
  \country{China}
}


\begin{abstract}
Talking head generation is a significant research topic that still faces numerous challenges. Previous works often adopt generative adversarial networks or regression models, which are plagued by generation quality and average facial shape problem. Although diffusion models show impressive generative ability, their exploration in talking head generation remains unsatisfactory. This is because they either solely use the diffusion model to obtain an intermediate representation and then employ another pre-trained renderer, or they overlook the feature decoupling of complex facial details, such as expressions, head poses and appearance textures. Therefore, we propose a \textbf{F}acial \textbf{D}ecoupled \textbf{D}iffusion model for \textbf{Talk}ing head generation called \textbf{FD2Talk}, which fully leverages the advantages of diffusion models and decouples the complex facial details through multi-stages. Specifically, we separate facial details into motion and appearance. In the initial phase, we design the Diffusion Transformer to accurately predict motion coefficients from raw audio. These motions are highly decoupled from appearance, making them easier for the network to learn compared to high-dimensional RGB images. Subsequently, in the second phase, we encode the reference image to capture appearance textures. The predicted facial and head motions and encoded appearance then serve as the conditions for the Diffusion UNet, guiding the frame generation. Benefiting from decoupling facial details and fully leveraging diffusion models, extensive experiments substantiate that our approach excels in enhancing image quality and generating more accurate and diverse results compared to previous state-of-the-art methods.
\end{abstract}




\maketitle

\section{Introduction}

Talking head generation is a task that creates a digital representation of a person's head and facial movements synchronized with the audio signal. This technology serves as a cornerstone with far-reaching applications, including virtual reality, augmented reality, and entertainment industries such as film production~\cite{zakharov2019few, zakharov2020fast, guo2021ad}. With the development of deep learning~\cite{yao2024multi, dosovitskiy2020image, yao2023ndc}, it has recently attracted numerous researchers and achieved impressive results.

Prevailing methodologies for talking head generation can be broadly divided into two paradigms. One approach involves using a GAN-based framework~\cite{doukas2021headgan, kr2019towards, gu2020flnet, das2020speech, prajwal2020lip, tewari2020stylerig}, which simultaneously optimizes a generator and a discriminator. However, due to the inherent flaws of GANs themselves and the suboptimal framework designs, this often results in unsatisfactory results, such as unnatural faces and inaccurate lip movements.
The other approach utilizes regression models~\cite{gururani2023space, zhou2020makelttalk, fan2022faceformer, lu2021live, wang2021audio2head, chen2019hierarchical} to map audio to facial movements, ensuring better temporal consistency. Nonetheless, regression-based methods encounter challenges in generating natural movements with individualized characteristics, leading to issues with average facial shapes and less diverse results.

\begin{figure}[t!]
  \includegraphics[width=\linewidth]{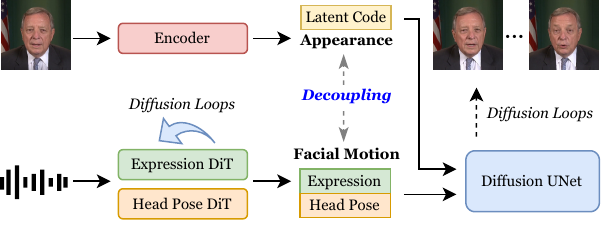}
  \caption{Our proposed FD2Talk leverages diffusion models to generate high-quality and diverse talking head videos. This framework decouples facial information into motion and appearance, thus maintaining motion plausibility, enhancing texture fidelity, and improving generalization.}
  \label{fig:intro}
\end{figure}

Recently, the rise of diffusion models~\cite{ho2020denoising, song2020denoising, rombach2022high} has marked a new era in generative tasks. Due to their stable generation process and relative ease of training, diffusion models offer a promising avenue for the advancement of talking head technology. While some previous works~\cite{shen2023difftalk, du2023dae, stypulkowski2024diffused} have attempted to apply diffusion models to talking head generation, their generated results still suffer from low image quality, unnaturalness, and insufficient lip synchronization. We analyze that there are two main issues in current methods. \textbf{1)} Some approaches~\cite{ma2023dreamtalk} apply diffusion models solely to predict facial intermediate representations, such as 3DMM coefficients. However, they still rely on pre-trained renderers for rendering the final faces, resulting in low-quality in the generated images. \textbf{2)} Other approaches~\cite{shen2023difftalk, stypulkowski2024diffused} directly generate faces through pixel-level denoising, globally conditioned on the audio and reference image. Nevertheless, they overlook the fact that faces contain rich information, such as expressions, poses, texture, etc. Previous methods couple these facial details, significantly complicating denoising generation and yielding unsatisfactory results.

To address the above issues, we propose the \textbf{F}acial \textbf{D}ecoupled \textbf{D}iffusion model for \textbf{Talk}ing head generation, named \textbf{FD2Talk}. Our FD2Talk leverages the generative advantages of diffusion models to generate high-quality, diverse and natural talking heads videos. As illustrated in~\cref{fig:intro}, the proposed FD2Talk model is a multi-stage framework that decouples complex facial details into motion and appearance information. The first phase focuses on motion information generation, while the second phase is dedicated to driving frame synthesis. \textbf{1) Motion Generation.} Motion information, including lip movements, expressions, and head poses, is highly related to the given audio and is more decoupled from facial appearance, making it easier to learn. In the first stage, we design novel Diffusion Transformers to extract motion-only information, \emph{i.e.}, 3DMM expression and head pose coefficients, from the raw audio. Through the denoising process, we generate natural and accurate motions, thereby enhancing the realism of our final outputs. Additionally, predicting the head pose coefficients at this stage enables us to produce more diverse motions compared to previous methods. \textbf{2) Frames Generation.} Moving on to the second stage, we first encode the reference image to capture appearance information, including human identity and texture characteristics. Combining this appearance information with the previously learned motion, we obtain a comprehensive facial representation related to the final RGB faces. Unlike previous methods that utilize a pre-trained face renderer to render final frames, we design a conditional Diffusion UNet and utilize motion and appearance as conditions to guide higher-quality and more natural animated frame generation. 

Our two-stage approach not only maintains motion plausibility and accuracy, but also enhances texture fidelity. \emph{Moreover, by focusing on generating appearance-independent information in the first stage, we can enhance the generalization ability of our FD2Talk.} This is because we can obtain pure motion coefficients from the audio signal without being influenced by the portrait domains. The contribution can be summarized as follows:

\begin{itemize}
    \item Our proposed FD2Talk is a multi-stage framework that effectively decouples facial motion and appearance, enabling accurate motion modeling, superior texture synthesis, and improved generalization.
    \item Our approach fully leverages the generative power of diffusion models in both motion and frames generation stages, thus enhancing the quality of the results.
    \item Extensive experiments demonstrate that our method excels at generating accurate and realistic talking head videos, achieving state-of-the-art performance. By incorporating head pose modeling, our FD2Talk produces significantly more diverse results compared to previous methods.
\end{itemize}
\section{Related Works}

\paragraph{Audio-Driven Talking Head Generation} Previous methods have attempted to utilize generative adversarial networks~\cite{prajwal2020lip, zhou2019talking, vougioukas2020realistic, chen2020talking, wang2020mead, chen2021talking, zhou2021pose, sun2021speech2talking} and regression models, such as RNN~\cite{suwajanakorn2017synthesizing}, LSTM~\cite{zhou2020makelttalk, gururani2023space, wang2021audio2head} and Transformer~\cite{fan2022faceformer, aneja2023facetalk} to synthesis talking head videos based on audio signals. Among GAN-based methods,~\cite{prajwal2020lip} proposed a novel lip-synchronization network that generates talking head videos with accurate lip movements across different identities by learning from a powerful lip-sync discriminator.~\cite{zhou2019talking} disentangled person identity and speech information through adversarial learning, leading to improved talking head generation.~\cite{vougioukas2020realistic} introduced a temporal GAN with three discriminators focused on achieving detailed frames, audio-visual synchronization, and realistic expressions, capable of generating lifelike talking head videos. On the other hand, in regression-based methods,~\cite{gururani2023space} adopts LSTM for better temporal consistency using explicit and implicit keypoints as the intermediate representation. Additionally,~\cite{fan2022faceformer} proposed a Transformer-based autoregressive model that encodes long-term audio context and autoregressively predicts a sequence of animated 3D face meshes. Despite significant progress, the unrealistic results in GAN-based generation and the average facial shape problem in regression-based models remain unresolved. 

\paragraph{Diffusion Models for Talking Head Generation} Diffusion models have demonstrated the remarkable ability across multiple generative tasks, such as image generation~\cite{saharia2022photorealistic, ramesh2022hierarchical, ruiz2023dreambooth}, image inpainting~\cite{lugmayr2022repaint, xie2023smartbrush, yang2023uni}, and video generation~\cite{ho2022imagen, blattmann2023stable, luo2023videofusion}. Recently, some studies~\cite{shen2023difftalk, du2023dae, stypulkowski2024diffused} have delved into using diffusion models for talking head generation. However, these studies still face challenges in producing natural and accurate faces. On one hand, they~\cite{ma2023dreamtalk} generate intermediate representations using diffusion models but rely on pre-trained face renderers for synthesizing the final frames. On the other hand, they~\cite{shen2023difftalk, stypulkowski2024diffused} globally utilize audio features to condition the generation of faces, which couples the complex facial motion and appearance. To fully leverage the advantages of the diffusion model and disentangle the complex facial information, we utilize the diffusion model in both motion generation and frame generation, thereby achieving better performance.
\section{Method}

Given a reference image $\mathcal{I}\in \mathbb{R}^{3\times H\times W}$ and a corresponding audio input, our model is designed to synthesize a realistic talking head video $\mathcal{V}\in \mathbb{R}^{3\times F\times H\times W}$ with lip movements synchronized with the audio signal. Here, the symbols $F$, $H$, and $W$ denote the frame numbers, frame height and frame width respectively.

Our FD2Talk framework consists of two stages that decouple facial information into motion and appearance, thus enhancing the modeling of facial representation. We employ powerful diffusion models in both stages, making FD2Talk a fully diffusion-based approach that produces high-quality talking head results. Specifically, we start by using Diffusion Transformers to predict expressions and pose motions from the audio input. In the subsequent stage, we utilize a Diffusion UNet to generate final RGB images, conditioned on the previously predicted motion information along with appearance texture information extracted from a reference image.

\subsection{Preliminary Knowledge}
\label{sec:preliminary_knowledge}

\subsubsection{3D Morphable Model} To generate high-quality talking heads, we integrate 3D information into our method, specifically employing the 3D Morphable Model (3DMM)~\cite{deng2019accurate} to decouple the facial representation from a given face image. This allows us to describe the 3D face space (3D mesh) using Principal Component Analysis:
\begin{equation}
\mathrm{\bf{S}}=\mathrm{\bf{S}}(\boldsymbol{\alpha}, \boldsymbol{\beta})=\Bar{\mathrm{\bf{S}}} + \mathrm{\bf{B}}_{id}\boldsymbol{\alpha} + \mathrm{\bf{B}}_{exp}\boldsymbol{\beta}.
\label{eq:3DMesh}
\end{equation}
Here, $\mathrm{\bf{S}}\in \mathbb{R}^{3N}$ (where $N$ represents the number of vertices of a face, and $3$ represents the axes $x$, $y$, and $z$) denotes a 3D face, while $\Bar{\mathrm{\bf{S}}}$ is the mean shape. $\boldsymbol{\alpha}\in \mathbb{R}^{D_{\alpha}}$ and $\boldsymbol{\beta}\in \mathbb{R}^{D_{\beta}}$ represent the predicted coefficients of identity and expression, respectively. $\mathrm{\bf{B}}_{id}$ and $\mathrm{\bf{B}}_{exp}$ are the PCA bases of identity and expression. Moreover, rotation coefficients $\boldsymbol{r}\in SO(3)$ and translation coefficients $\boldsymbol{t}\in \mathbb{R}^{3}$ represent the head rotation and translation, respectively, collectively constituting the facial pose coefficients $\boldsymbol{p}=[\boldsymbol{r}, \boldsymbol{t}]$.

\subsubsection{Diffusion Model} Diffusion models are formulated as time-conditional denoising networks that learn the reverse process of a Markov Chain with a length $T$. Specifically, starting from the clean signal $\boldsymbol{x}_{0}$, the process of adding noise can be denoted as follows:
\begin{equation}
\boldsymbol{x}_t = \sqrt{\overline{\alpha}_t}\boldsymbol{x}_0 + \sqrt{1-\overline{\alpha}_t}\epsilon_t.
\end{equation}
Here, $\epsilon_t\sim \mathcal{N}(0,1)$ denotes random Gaussian noise, while $\overline{\alpha}_t$ represents the hyper-parameter for the diffusion process. $\boldsymbol{x}_t$ refers to the noisy feature at step $t$, where $t\in [1, \ldots, T]$. During inference, the $T$-step denoising process progressively denoise random Gaussian noise $\mathcal{N}(0,1)$ to estimate the clean signal $\boldsymbol{x}_{0}$. In our work, all diffusion-based models are designed to predict signal itself rather than noise. Thus, the overall goal can be described as follows:
\begin{equation}
L:=\mathbb{E}_{\boldsymbol{x}_0, t}\left[\left\|\boldsymbol{x}_0-\theta(\boldsymbol{x}_t, t, \boldsymbol{c})\right\|_{2}^{2}\right],
\end{equation}
where $\theta$ represents the diffusion model and $c$ represents conditional guiding. We utilize the $L2$ error between the estimated signal and the ground truth $\boldsymbol{x}_0$.

\begin{figure}[t!]
  \includegraphics[width=\linewidth]{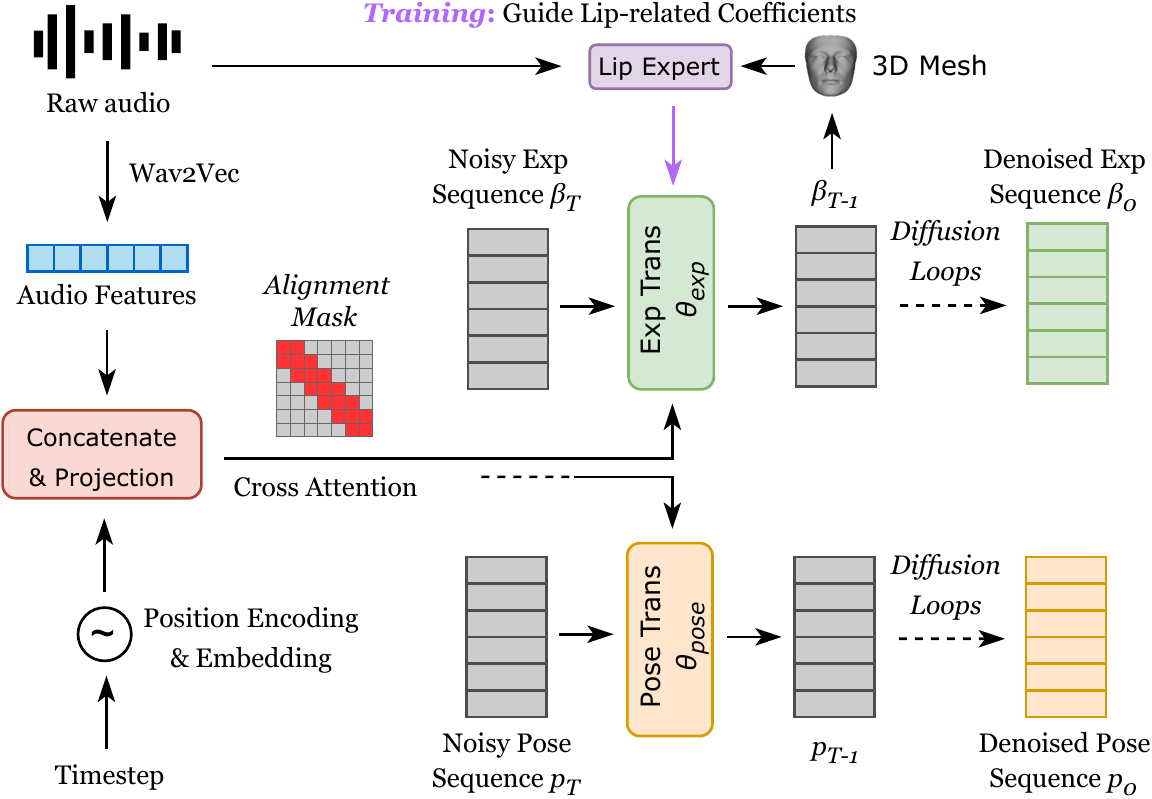}
  \caption{Pipeline of the motion generation. We decouple the motion into expression and head poses, both of which are predicted by our designed DiTs. The audio guides the generation through cross-attention layers, utilizing an alignment mask to ensure accurate lip movements. Furthermore, the pre-trained lip expert also enhances the lip synchronization.}
  \label{fig:stage1}
\end{figure}

\subsection{Motion Generation with Diffusion Transformers}

Early diffusion-based methods~\cite{shen2023difftalk, stypulkowski2024diffused} globally utilize audio signals as a condition for the pixel-level denoising process. However, this approach combines motion and appearance, making it challenging for overall training convergence. In contrast, in the first stage, our method focuses on generating motion-only information from the audio signal, specifically 3DMM expression and head pose coefficients. These coefficients exclusively represent facial and head motion, which are highly decoupled from the appearance textures and greatly influence lip synchronization and motion diversity. Furthermore, compared to high-dimensional RGB faces, low-dimensional 3DMM coefficients are considerably easier for the model to learn.

To ensure smooth continuity between different frame motions and fully leverage the diffusion models, we introduce sequence-to-sequence Diffusion Transformers for generating both expression and pose coefficients. Meanwhile, to effectively address the one-to-many mapping problem and accurately predict lip movements and diverse head poses, we decouple the prediction of expression and pose coefficients using an Expression Transformer $\theta_{exp}$ and a Pose Transformer $\theta_{pose}$, which is illustrated in~\cref{fig:stage1}. 

Specifically, we initialize the noisy expression sequence $\boldsymbol{\beta}_T\in \mathbb{R}^{F\times D_{\beta}}$ and noisy pose sequence $\boldsymbol{p}_T\in \mathbb{R}^{F\times D_{p}}$ from random Gaussian noise $\mathcal{N}(0,1)$, where $F$ represents the number of frames aligned with the final video. We then denoise the $\boldsymbol{\beta}_T$ and $\boldsymbol{p}_T$ conditioned on audio features through $T$ loops to estimate denoised sequence $\boldsymbol{\beta}_0$ and $\boldsymbol{p}_0$. Here, the length of audio clip $\boldsymbol{A}$ is aligned with $F$, and we adopt the state-of-the-art self-supervised pre-trained speech model, Wav2Vec 2.0~\cite{baevski2020wav2vec}, to extract the audio features. 

Taking $\theta_{pose}$ as an example, at each timestep $t$, we concatenate the embedding from the timestep and audio features to obtain the condition $c$. We then project $c$ to an intermediate representation $\tau(c)\in \mathbb{R}^{F\times D_{\tau}}$ using a linear layer. Then, $\tau(c)$ is fused into $\theta_{pose}$ via the cross-attention layer, where the query ($\mathrm{\bf{Q}}$) is derived from $\boldsymbol{p}_t$, while the key ($\mathrm{\bf{K}}$) and value ($\mathrm{\bf{V}}$) are obtained from $\tau(c)$. Meanwhile, we design an alignment mask $\mathcal{M}$ to ensure the consistency of generated coefficients and the audio signal, so that $\tau(c)$ for the $i^{th}$ timestamp attends to $\boldsymbol{p}_t$ at the $j^{th}$ timestamp only if $j-k\leq i\leq j+k$. For FD2Talk, we empirically set $k = 3$. The $\mathcal{M}$ can be denoted as:
\begin{equation}
\mathcal{M}=\left \{
\begin{array}{ll}
    True, & \text{if } j-k\leq i\leq j+k \\
    False, & \text{otherwise}
\end{array}
\right.
\end{equation}

In our diffusion process, we directly estimate the original signal. Therefore, after $L$-layer Pose Transformer, we obtain $\boldsymbol{\Tilde{p}_{0}}$. Subsequently, we can calculate the single-step denoising result $\boldsymbol{p}_{t-1}$:
\begin{equation}
\begin{split}
\boldsymbol{p}_{t-1}=\sqrt{\overline{\alpha}_{t-1}}\boldsymbol{\Tilde{p}_{0}}+\frac{\sqrt{1-\overline{a}_{t-1}-\sigma_t^2}}{\sqrt{1-\overline{a}_t}}(\boldsymbol{p}_{t}-\sqrt{\overline{a}_t}\boldsymbol{\Tilde{p}_{0}})+\sigma_t\epsilon,
\label{eq:denoise}
\end{split}
\end{equation}
where $\sigma_t$ is the Gaussian covariance at the $t^{th}$ timestep.

The Exp Transformer $\theta_{exp}$ and Pose Transformer $\theta_{pose}$ share the same architecture, and the denoising process for $\boldsymbol{\beta}_t$ is identical to that for $\boldsymbol{p}_t$. Therefore, after $T$ iterations, we obtain $\boldsymbol{\beta}_0$ and $\boldsymbol{p}_0$ as the final values for the expression and pose coefficients.

\subsection{Frame Generation with Diffusion UNet}

Previous methods primarily employ pre-trained face renderers~\cite{ren2021pirenderer, wang2021one} to generate final RGB faces, whose performance sets an upper bound on talking face generation. Therefore, we design a conditional Diffusion UNet $\theta_{unet}$ to generate the final frame based on previously predicted 3DMM coefficients, aiming to utilize the diffusion models to achieve diverse and realistic faces generation.

To reduces computational overhead and accelerates convergence, we introduce a pair of encoder $\mathcal{E}$ and decoder $\mathcal{D}$~\cite{rombach2022high} to transition the frame generation into the latent space. Suppose the downsampling factor is $f=H/h=W/w$, then we can encode the reference image $\mathcal{I}$ into the reference latent code $x=\mathcal{E}(\mathcal{I})\in \mathbb{R}^{d\times h\times w}$.

\begin{figure}[t!]
  \includegraphics[width=\linewidth]{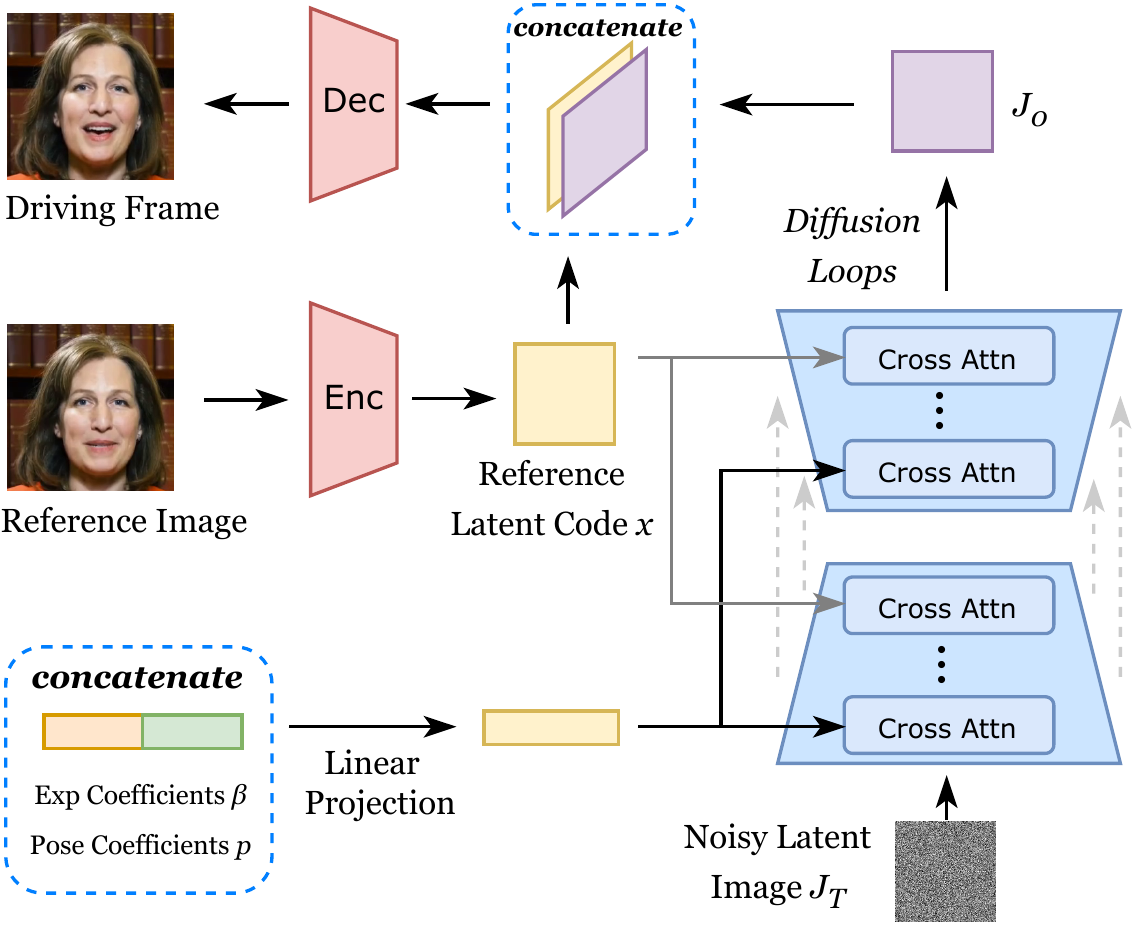}
  \caption{Pipeline of the frame generation. The facial appearance extracted from the reference image and the predicted motion coefficients are fused within the Diffusion UNet using distinct cross-attention layers to prevent interference.}
  \label{fig:stage2}
\end{figure}

As shown in~\cref{fig:stage2}, we initialize the noisy latent image $\boldsymbol{J}_T\in \mathbb{R}^{d\times h\times w}$ from $\mathcal{N}(0,1)$, then we progressively denoise it conditioned on both reference latent code $x$ and 3DMM coefficients $\boldsymbol{\beta}_0$ and $\boldsymbol{p}_0$. Here, $x$ encompasses the appearance texture of the reference image, while $\boldsymbol{\beta}_0$ and $\boldsymbol{p}_0$ includes the driving facial and head motions. 

An intuitive approach is to directly concatenate the $x$, $\boldsymbol{\beta}_0$ and $\boldsymbol{p}_0$ to obtain the conditions. However, we observe that this operation leads to difficulties in training convergence, because there exists a gap between the image domain and the motion coefficients domain. To address the impact of domain gap, we use two cross-attention layers to introduce these two conditions respectively. Specifically, both the encoder and decoder of Diffusion UNet consist of two cross-attention layers, denoted as $\phi_1$ and $\phi_2$. The coefficients $\boldsymbol{\beta}_0$ and $\boldsymbol{p}_0$ are concatenated, following with a linear projection, to form the condition for $\phi_1$. The calculation of $\phi_1$ can be defined as:
\begin{equation}
\boldsymbol{m}_1=\phi_1(\{\boldsymbol{\beta}_0, \boldsymbol{p}_0\}, \boldsymbol{J}_t),
\end{equation}
where the query ($\mathrm{\bf{Q}}$) is from $\boldsymbol{J}_t$, and the key ($\mathrm{\bf{K}}$) and value ($\mathrm{\bf{V}}$) are from the condition $\{\boldsymbol{\beta}_0, \boldsymbol{p}_0\}$. Then, in the second layer $\phi_2$, we utilize the reference latent code $x$ as the condition to guide this process:
\begin{equation}
\boldsymbol{m}_2=\phi_2(x, \boldsymbol{m}_1),
\end{equation}
where the query ($\mathrm{\bf{Q}}$) is from $\boldsymbol{m}_1$, and the key ($\mathrm{\bf{K}}$) and value ($\mathrm{\bf{V}}$) are derived from $x$. Here, the $x$ is reshaped into sequence, and positional encoding is also introduced. This decoupling of conditions enhances the denoising stability, leading to higher-quality results.

Similar to that in the first stage, at each diffusion timestep $t$, we predict $\boldsymbol{\Tilde{J}}_0$ from $\boldsymbol{J}_t$, and then calculate the corresponding $\boldsymbol{J}_{t-1}$ using the~\cref{eq:denoise}. After $T$ iterations, this process generates the accurate denoised latent image $\boldsymbol{J}_0$. The reference latent code and the denoised latent image are further concatenated as the input of decoder $\mathcal{D}$, allowing us to generate the RGB image $\mathcal{V}_i$, which serves as each frame for the talking head video $\mathcal{V}=\{\mathcal{V}_i\}^F_1$. Moreover, as we denoise in the latent space, we can easily extend to higher-resolution talking head synthesis by adjusting the downsampling factor $f$, thereby further enhancing our generation quality.

\begin{table*}[t!]
    \caption{Comparison with the state-of-the-art methods on HDTF and VoxCeleb dataset. The best results are highlighted in bold, and the second best is underlined. Our FD2Talk surpasses previous methods in motion diversity and image quality, as well as offering competitive lip synchronization performance. The data presented in the table are in the order of \emph{HDTF / VoxCeleb}.}
    \label{tab:main_results}
    \centering
    \begin{tabular}{l|cc|cc|ccc}
    \hline
    \multirow{2}{*}{Methods} & \multicolumn{2}{c|}{Lip Synchronization} & \multicolumn{2}{c|}{Motion Diversity} & \multicolumn{3}{c}{Image Quality} \\
    & LSE-C $\uparrow$ & SyncNet $\uparrow$ & Diversity $\uparrow$ & Beat Align $\uparrow$ & FID $\downarrow$ & PSNR $\uparrow$ & SSIM $\uparrow$ \\
    \hline
    Ground Truth & 8.32 / 6.29 & 7.99 / 5.73 & 0.256 / 0.307 & 0.276 / 0.319 & --- & --- & --- \\
    \hline
    Wav2Lip~\cite{prajwal2020lip} & {\bf 10.08} / {\bf 8.13} & {\bf 8.06} / {\bf 6.40} & N./A. / N./A. & N./A. / N./A. & \underline{22.67} / 23.85 & 32.33 / 35.19 & 0.740 / 0.653 \\
    MakeItTalk~\cite{zhou2020makelttalk} & 4.89 / 2.96 & 3.72 / 2.67 & 0.238 / 0.260 & 0.221 / 0.252 & 28.96 / 31.77 & 17.95 / 21.08 & 0.623 / 0.529 \\
    SadTalker~\cite{zhang2023sadtalker} & 6.11 / 4.51 & 5.19 / 4.88 & \underline{0.275} / \underline{0.319} & \underline{0.296} / \underline{0.328} & 23.76 / 24.19 & 35.78 / \underline{37.90} & \underline{0.746} / 0.690 \\
    DiffTalk~\cite{shen2023difftalk} & 6.06 / 4.38 & 4.98 / 4.67 & 0.235 / 0.258 & 0.226 / 0.253 & 23.99 / 24.06 & \underline{36.51} / 36.17 & 0.721 / 0.686 \\
    DreamTalk~\cite{ma2023dreamtalk} & 6.93 / 4.76 & 5.46 / 4.90 & 0.236 / 0.257 & 0.213 / 0.249 & 24.30 / \underline{23.61} & 32.82 / 33.16 & 0.738 / \underline{0.692} \\
    \hline
    Ours & \underline{7.29} / \underline{5.16} & \underline{6.63} / \underline{5.66} & {\bf 0.338} / {\bf 0.359} & {\bf 0.336} / {\bf 0.377} & {\bf 20.96} / {\bf 21.89} & {\bf 38.89} / {\bf 39.95} & {\bf 0.779} / {\bf 0.756} \\
    \hline
    \end{tabular}
\end{table*}

\subsection{Training Strategies}

Our training process consists of two stages. In the first stage, we train the Exp Transformer and Pose Transformer to generate accurate expression and pose coefficients. Using these accurate coefficients as a foundation, we then train the Diffusion UNet in the second stage to generate natural and diverse RGB frames.

\subsubsection{Motion Generation Stage} 

In the first stage, we randomly extract a video clip along with the corresponding audio clip $\boldsymbol{A}$ from the training set. We utilize the Deep3d~\cite{deng2019accurate} method to generate the expression coefficient sequence $\boldsymbol{\beta}_0$ and pose coefficient sequence $\boldsymbol{p}_0$ from this video clip. $\boldsymbol{\beta}_0$ and $\boldsymbol{p}_0$ also serve as the ground truths. Then, our Exp Transformer $\theta_{exp}$ and Pose Transformer $\theta_{pose}$ can be trained using the tuples $(\boldsymbol{\beta}_0, t, A)$ and $(\boldsymbol{p}_0, t, A)$, respectively.

For the Exp Transformer $\theta_{exp}$, by adding random Gaussian noise, the $\boldsymbol{\beta}_0$ can become $\boldsymbol{\beta}_t$ using~\cref{eq:denoise}. The $\theta_{exp}$ estimates $\boldsymbol{\Tilde{\beta}_{0}}=\theta_{exp}(\boldsymbol{\beta}_t, t, A)$, and the objective can be defined as follows:
\begin{equation}
\mathcal{L}_{exp}=\mathbb{E}_{\boldsymbol{\beta}_0, t}\left[\left\|\boldsymbol{\beta}_0-\theta_{exp}(\boldsymbol{\beta}_t, t, A)\right\|_{2}^{2}\right].
\end{equation}
Similar to the $\theta_{exp}$, the objective of the Pose Transformer $\theta_{pose}$ is:
\begin{equation}
\mathcal{L}_{pose}=\mathbb{E}_{\boldsymbol{p}_0, t}\left[\left\|\boldsymbol{p}_0-\theta_{pose}(\boldsymbol{p}_t, t, A)\right\|_{2}^{2}\right].
\end{equation}

While the random noise introduced in the diffusion model can effectively facilitate the diverse generation, it also leads to inaccurate mouth shape generation to some extent. Therefore, we utilize a pre-trained lip expert~\cite{prajwal2020lip} to guide this denoising process and generate more accurate mouth shape. Specifically, we first obtain the identity coefficients from the reference image, and then calculate the 3D meshes using these identity coefficients along with the predicted expression coefficients $\boldsymbol{\Tilde{\beta}_{0}}$ via~\cref{eq:3DMesh}. From these 3D meshes, we select vertices in the mouth area to represent lip motion~\cite{ma2023styletalk}. The pre-trained lip expert calculates the cosine similarity between mouth motion embedding $v$ and audio embedding $a$ as follows:
\begin{equation}
P_{sync}=\frac{v\times a}{max(\|v\|_2\times \|a\|_2, \epsilon)},
\end{equation}
where $\epsilon$ is a small number for avoiding the division-by-zero. Then, the $\theta_{exp}$ minimizes the synchronous loss as follows:
\begin{equation}
\mathcal{L}_{sync}=-\mathrm{log}(P_{sync}).
\end{equation}

Overall, the first stage optimizes the following loss:
\begin{equation}
\mathcal{L}_{first}=\lambda_{exp}\mathcal{L}_{exp}+\lambda_{pose}\mathcal{L}_{pose}+\lambda_{sync}\mathcal{L}_{sync},
\end{equation}
where $\lambda_{exp}$, $\lambda_{pose}$ and $\lambda_{sync}$ are the weight factors to control the three losses in the same numeric scale.

\subsubsection{Frame Generation Stage}

We utilize the pre-trained~\cite{esser2021taming} encoder $\mathcal{E}$ and decoder $\mathcal{D}$ as the foundation for learning in the latent space. Given that the input channel for the decoder in our method is $2\times d$, we opt to substitute the first convolution layer of the decoder. Subsequently, we fine-tune both the encoder and decoder using frames from the training set. Specifically, in each iteration, we randomly select two frames $F_1$ and $F_2$ from a single video and then calculate the reconstruction loss as follows:
\begin{equation}
\mathcal{L}_{rec}=\left\|F_2-\mathcal{D}([\mathcal{E}(F_1), \mathcal{E}(F_2)])\right\|_{2}^{2}.
\end{equation}
Meanwhile, we introduce the perceptual loss~\cite{zhang2018unreasonable} to enforce $\mathcal{E}$ and $\mathcal{D}$ to accurately reconstruct the frames in the image space:
\begin{equation}
\mathcal{L}_{per}=\left\|\phi(F_2)-\phi(\mathcal{D}([\mathcal{E}(F_1), \mathcal{E}(F_2)]))\right\|_{1},
\end{equation}
where $\phi$ represents the perceptual feature extractor~\cite{zhang2018unreasonable}. Then the overall objective of encoder $\mathcal{E}$ and decoder $\mathcal{D}$ can be defined as:
\begin{equation}
\mathcal{L}_{e\&d}=\lambda_{rec}\mathcal{L}_{rec}+\lambda_{per}\mathcal{L}_{per},
\end{equation}
where $\lambda_{rec}$ and $\lambda_{per}$ control the numeric scales.

\begin{figure*}[t!]
  \includegraphics[width=\linewidth]{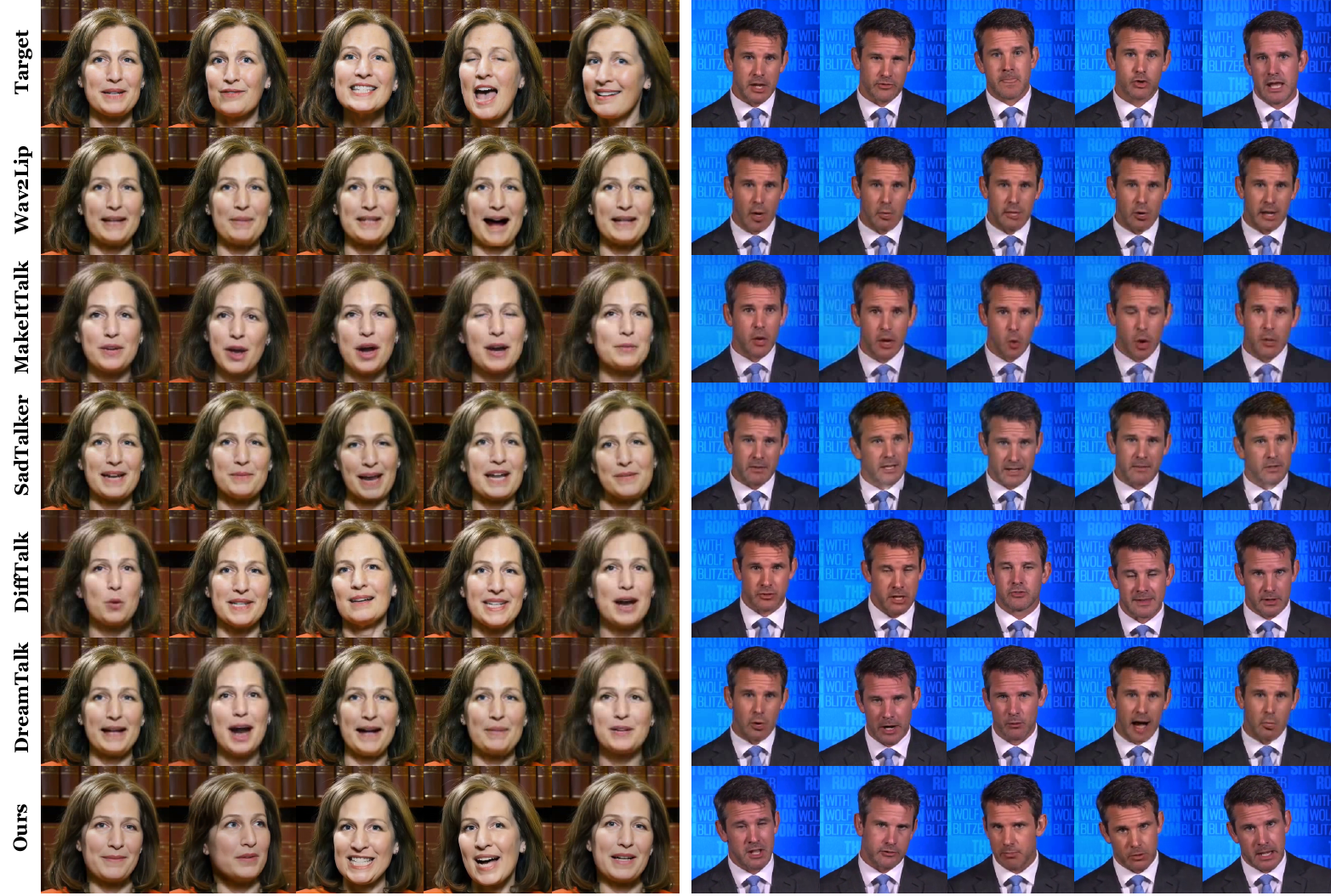}
  \caption{Qualitative comparison with several state-of-the-art methods. Our FD2Talk achieves superior lip synchronization compared to previous methods while preserving naturalness and high image quality. By leveraging diffusion models for predicting head motion, our generated results also exhibit enhanced motion diversity.}
  \label{fig:results}
\end{figure*}

During the training of Diffusion UNet $\theta_{unet}$, we randomly extract a video clip along with its corresponding audio clip. The first frame from this video clip serves as the reference image $\mathcal{I}$. Utilizing the trained encoder $\mathcal{E}$, we obtain the reference latent code $x$, as well as the ground truths for each latent image $\boldsymbol{J}_0$. Subsequently, we employ the trained Exp Transformer and Pose Transformer to acquire the $\boldsymbol{\Tilde{\beta}}_0\in \mathbb{R}^{F\times D_{\beta}}$ and $\boldsymbol{\Tilde{p}}_0\in \mathbb{R}^{F\times D_p}$. Different from the sequence-to-sequence Diffusion Transformer in the first stage, our Diffusion UNet generate each RGB frames one by one, so we extract the coefficient $\boldsymbol{\Tilde{\beta}}\in \mathbb{R}^{D_{\beta}}$ and $\boldsymbol{\Tilde{p}}\in \mathbb{R}^{D_p}$ for each frame. The training of our Diffusion UNet is facilitated by a tuple denoted as $(\boldsymbol{J}_0, t, \boldsymbol{\Tilde{\beta}}, \boldsymbol{\Tilde{p}}, x)$. Specifically, we add the random Gaussian noise on $\boldsymbol{J}_0$ to obtain the noisy latent image $\boldsymbol{J}_t$ at the $t$-th timestep. We then optimize $\theta_{unet}$ using the following objective function:
\begin{equation}
\mathcal{L}_{second}=\mathbb{E}_{\boldsymbol{J}_0, t}\left[\left\|\boldsymbol{J}_0-\theta_{unet}(\boldsymbol{J}_t, t, \boldsymbol{\Tilde{\beta}}, \boldsymbol{\Tilde{p}}, x)\right\|_{2}^{2}\right].
\end{equation}
\section{Experiments}

\subsection{Experimental Setup}

\paragraph{Datasets.} We use HDTF~\cite{zhang2021flow} and VFHQ~\cite{xie2022vfhq} datasets to train our FD2Talk. HDTF is a large in-the-wild high-resolution and high-quality audio-visual dataset that consists of about 362 different videos spanning 15.8 hours. The resolution of the face region in the video generally reaches $512 \times 512$. VFHQ is a large-scale video face dataset, which contains over 16000 high-fidelity clips of diverse interview scenarios. However, since VFHQ lacks audio components, it is exclusively utilized during the second phase of training. All videos are clipped into small fragments and cropped~\cite{siarohin2019first} to obtain the face region. Then we use Deep3d~\cite{deng2019accurate}, a single-image face reconstruction method, to recover the facial image and extract the relevant coefficients. Both HDTF and VFHQ are split $70\%$ as the training set, $10\%$ as the validation set, and $20\%$ as the testing set. Moreover, we introduce VoxCeleb~\cite{nagrani2017voxceleb} to further evaluate our method, which contains over 100$k$ videos of 1251 subjects.

\paragraph{Implementation Detail.} We train the model on video frames with $256 \times 256$ resolution. In the first stage, the $6$-layer Exp Transformer and Pose Transformer are trained with a batch size of 1 and a generated sequence length of 25. In the second stage, we first fine-tune the pre-trained~\cite{esser2021taming} encoder and decoder, and then train the Diffusion UNet with a batch size of 32, and the resolution of the latent image is $64 \times 64$. The two-stage framework is trained with the Adam~\cite{da2014method} optimizer separately and can be inferred in an end-to-end fashion. The diffusion step is set to 1000 and 50 during training and inference, respectively. Our two-stage model is trained for approximately 8 and 32 hours using 8 NVIDIA 3090 GPUs.

\paragraph{Baselines.} We compare our method with several previous methods of audio-driven talking head generation, including Wav2Lip~\cite{prajwal2020lip}, MakeItTalk~\cite{zhou2020makelttalk}, SadTalker~\cite{zhang2023sadtalker}, DiffTalk~\cite{shen2023difftalk}, and DreamTalk~\cite{ma2023dreamtalk}. We provide a reference image and audio signal as input for all methods. Note that Wav2Lip requires additional videos to offer head pose information, so we also fixed the head pose of our method for a fair comparison in quantitative evaluation. 

\paragraph{Evaluation Metrics.} To evaluate the superiority of our proposed method, we consider three aspects: 1) Lip synchronization is assessed using two metrics: LSE-C~\cite{prajwal2020lip} and SyncNet~\cite{chung2017out}. LSE-C measures the confidence score of perceptual differences in mouth shape from Wav2Lip, while the SyncNet score assesses the audio-visual synchronization quality. 2) Motion diversity is evaluated by extracting head motion feature embeddings using Hopenet~\cite{ruiz2018fine} and calculating their standard deviations. Additionally, we use the Beat Align Score~\cite{siyao2022bailando} to measure alignment between the audio and generated head motions. 3) Generated image quality is assessed using widely recognized metrics: FID~\cite{heusel2017gans}, PSNR, and SSIM~\cite{wang2004image}.

\subsection{Qualitative Comparison}
We compare our method with previous state-of-the-art methods qualitatively. The results are visualized in~\cref{fig:results}. While Wav2Lip can generate accurate lip movements, it falls short in producing high-quality images due to blurriness issues in the mouth region. Moreover, Wav2Lip focuses solely on animating the lips, neglecting other facial areas and resulting in a lack of motion diversity. MakeItTalk and SadTalker attempt to address some weaknesses of Wav2Lip, such as enhancing motion diversity. However, they still struggle to synthesize detailed facial features like apple cheeks and teeth due to generative limitations in GANs and regression models. For diffusion-based methods, DiffTalk combines appearance and motion during denoising, leading to inaccurate lip movement generation. DreamTalk, on the other hand, neglects head pose modeling and still relies on pre-trained render models, resulting in synthesized results with unreasonable head poses and slightly distorted facial regions. In contrast, our FD2Talk fully leverages powerful diffusion models in both stages and effectively separates appearance and motion information. These operations result in accurate lip movements, diverse head poses, and high-quality, lifelike talking head videos.

\begin{figure}[t!]
  \includegraphics[width=\linewidth]{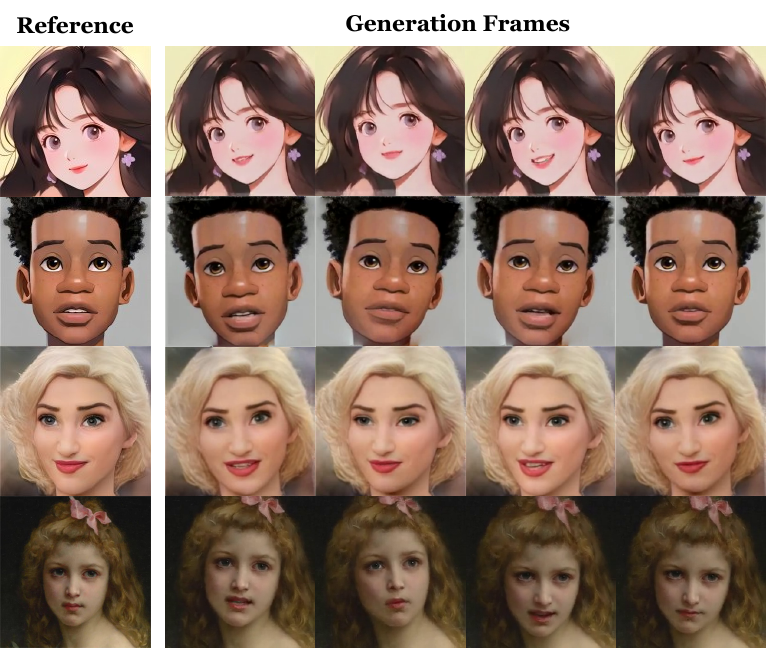}
  \caption{Our FD2Talk demonstrates strong generalization when applied to out-of-domain portraits. We generate each talking head video using the same audio but different portrait domains, which significantly diverge from training data.}
  \label{fig:generalization}
\end{figure}

\subsection{Quantitative Comparison}
We further quantitatively analyze the comparison between FD2Talk and previous state-of-the-art methods in lip synchronization, motion diversity, and image quality, on HDTF and VoxCeleb datasets. 

Our approach surpasses MakeItTalk, SadTalker, DiffTalk, and DreamTalk in terms of lip synchronization. We attribute this improvement to the alignment mask used during cross-attention in the Exp and Pose Transformer. This mask enables the predicted coefficients consistency with the corresponding audio signal. Additionally, the accurate lip movements are further enhanced by the lip synchronization loss with a well-pretrained lip expert. It is worth noting that although Wav2Lip achieves the highest lip accuracy, it neglects the overall naturalness and diversity of the results.

When considering the three metrics of image quality, \emph{i.e.}, FID, PSNR, and SSIM, our approach significantly outperforms previous methods, which can be attributed to two aspects: 1) Our method maximizes the potential of diffusion models to generate more natural results compared to previous works using GANs, regression models, or partial diffusion models. 2) We disentangle complex facial information through two stages, enabling accurate motion prediction, and the creation of natural, high-fidelity appearance textures, ultimately resulting in superior and high-quality results.

Moreover, our work surpasses previous methods in the diversity of head motions and achieves the best performance in Diversity and Beat Align Score. This achievement is attributed to our Pose Transformer, which predicts the head pose coefficients through the denoising process. The introduced random noise facilitates the generation of richer and more diverse pose results compared to previous methods.

\subsection{Generalization Performance}

We also test the generalization of our FD2Talk model for out-of-domain portraits. As demonstrated in~\cref{fig:generalization}, whether the provided faces are paintings, cartoon portraits, or oil paintings, our FD2Talk can animate them using audio signals, ensuring lip synchronization while preserving the appearance details of the reference face with high fidelity, thus enhancing image quality. Moreover, the generated results include rich head poses, demonstrating excellent motion diversity as expected. This generalization ability stems from the decoupling of facial representation. In the first stage, we focus on generating appearance-independent motion information, which is solely linked to the audio signal and remains robust across various portrait domains.

\subsection{Ablation Studies}

\begin{figure}[t!]
  \includegraphics[width=\linewidth]{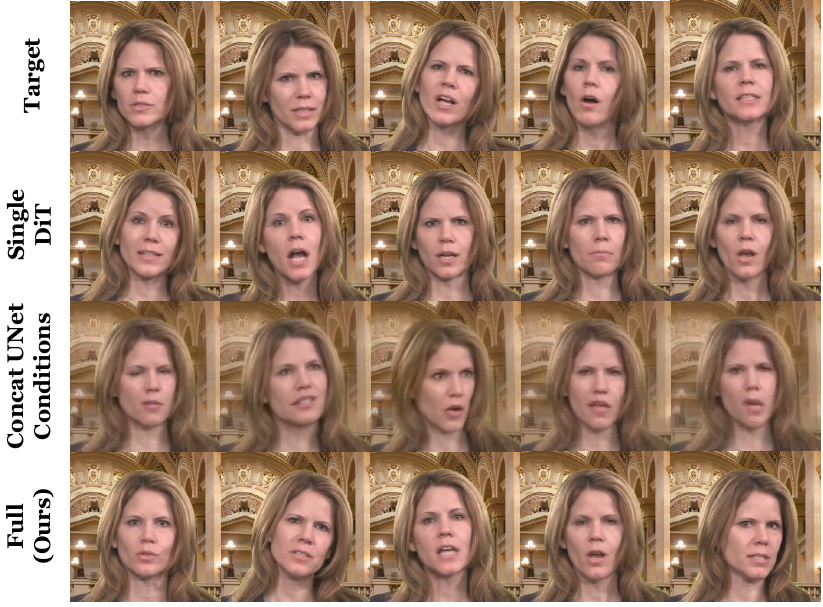}
  \caption{The visualization results of: 1) Utilizing a single DiT to predict expressions and head poses jointly; 2) Concatenating the two conditions of UNet; and 3) Our full FD2Talk model. We can observe that using a single DiT makes the results less diverse and synchronized, while concatenating two conditions leads to distorted and unnatural faces.}
  \label{fig:ablation}
\end{figure}

\begin{table*}[t!]
    \caption{Ablation studies on the 1) Decoupling of Diffusion Transformers and 2) Conditions of the Diffusion UNet.}
    \label{tab:ablation}
    \centering
    \begin{tabular}{l|cc|cc|ccc}
    \hline
    \multirow{2}{*}{Settings} & \multicolumn{2}{c|}{Lip Synchronization} & \multicolumn{2}{c|}{Motion Diversity} & \multicolumn{3}{c}{Image Quality} \\
    & LSE-C $\uparrow$ & SyncNet $\uparrow$ & Diversity $\uparrow$ & Beat Align $\uparrow$ & FID $\downarrow$ & PSNR $\uparrow$ & SSIM $\uparrow$ \\
    \hline
    Single Diffusion Transformer & 3.11 & 3.08 & 0.193 & 0.189 & 22.06 & 37.66 & 0.761 \\
    Concatenate UNet Conditions & 4.79 & 4.66 & 0.249 & 0.251 & 30.79 & 29.91 & 0.523 \\
    Full (Our FD2Talk) & {\bf 7.31} & {\bf 6.26} & {\bf 0.322} & {\bf 0.331} & {\bf 21.32} & {\bf 38.10} & {\bf 0.776} \\
    \hline
    \end{tabular}
\end{table*}

\subsubsection{Decoupling Expressions and Head Poses} In the first stage, we decouple the Diffusion Transformers for the prediction of expressions and poses to address the one-to-many mapping issue. We compare this approach with a baseline where a Diffusion Transformer is used to jointly predict expression and pose coefficients. As shown in~\cref{tab:ablation} and~\cref{fig:ablation}, this baseline exhibits a noticeable decrease in lip synchronization and motion diversity. This is because lip movements are heavily influenced by facial expressions but have little correlation with head pose. On the other hand, motion diversity is closely related to predicted pose coefficients. Jointly learning these coefficients leads to mutual interference and makes training more challenging. Therefore, we choose to decouple the prediction of expression and pose coefficients using Exp and Pose Transformers, respectively.

\subsubsection{Conditions of Diffusion UNet} In the second stage, the predicted motion information and encoded appearance texture are passed through distinct cross-attention layers to guide the Diffusion UNet. We verify its effectiveness by comparing it with a baseline where we directly concatenate these two conditions and guide the denoising process. As demonstrated in~\cref{tab:ablation} and~\cref{fig:ablation}, concatenating motion and appearance leads to a decrease in each metric, particularly image quality, as we can observe the distortion in faces. We analyze that appearance textures constitute image-domain information, which is much higher than coefficient-domain motion. Therefore, decoupling them using two distinct cross-attention layers can significantly enhance the robustness of overall diffusion models and ensure convergence.

\subsubsection{Ablation Studies of Lip Synchronization} 
\begin{figure}[t!]
  \includegraphics[width=\linewidth]{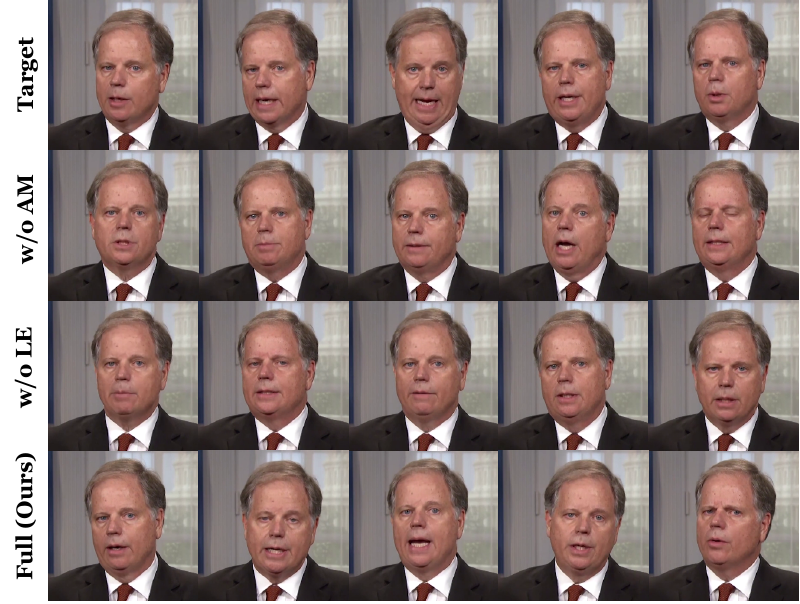}
  \caption{Comparison of 1) w/o AM: without alignment mask, 2) w/o LE: training without lip expert, and 3) full FD2Talk. We can notice that both the alignment mask and pre-trained lip expert can enhance lip synchronization of our model.}
  \label{fig:lip}
\end{figure}

In FD2Talk, we ensure lip synchronization from two aspects: \textbf{1) Aligning the audio and motions during cross-attention.} When we integrate audio features into the network, an alignment mask $\mathcal{M}$ is designed to ensure the consistency of generated coefficients and audio. To assess its significance, we conduct an experiment by removing the $\mathcal{M}$. As indicated in~\cref{tab:lip}, the absence of $\mathcal{M}$ notably affects lip synchronization. Our analysis demonstrates that without $\mathcal{M}$, the motion generation in each timestamp is misled by audio from other unrelated timestamps. \textbf{2) Guided with the pre-trained lip expert.} During the training of Exp Transformer, we utilize a pre-trained lip expert to constrain the lip-related coefficients using $\mathcal{L}_{sync}$. Here, we remove it to compare the effectiveness of $\mathcal{L}_{sync}$. As shown in~\cref{tab:lip}, when the model is trained without $\mathcal{L}_{sync}$, lip synchronization significantly drops. We attribute this to the fact that the coefficients are generated through a denoising process, which means introduced random noise may lead to inaccurate lip shapes.~\cref{fig:lip} also shows that utilizing the alignment mask and training with $\mathcal{L}_{sync}$ result in much better lip synchronization for the generated faces.

\begin{table}[t!]
    \caption{Ablation studies of lip synchronization. w/o alignment: We remove the alignment mask in DiTs. w/o $\mathcal{L}_{sync}$: We eliminate the constraint from the pre-trained lip expert.}
    \label{tab:lip}
    \centering
    \begin{tabular}{l|cc}
    \hline
    \multirow{2}{*}{Settings} & \multicolumn{2}{c}{Lip Synchronization} \\
    & LSE-C $\uparrow$ & SyncNet $\uparrow$ \\
    \hline
    w/o alignment & 4.66 & 3.97 \\
    w/o $\mathcal{L}_{sync}$ & 5.35 & 4.63 \\
    Full (Our FD2Talk) & {\bf 7.31} & {\bf 6.26} \\
    \hline
    \end{tabular}
\end{table}

\subsection{User Studies}

\begin{table}[t!]
    \caption{User studies results.}
    \label{tab:user_studies}
    \centering
    \begin{tabular}{l|c|c|c}
    \hline
    Methods & Lip Sync & Motion Diversity & Image Quality \\
    \hline
    Wav2Lip & \underline{24.9\%} & 1.2\% & 2.1\% \\
    MakeItTalk & 3.6\% & 2.7\% & 3.5\% \\
    SadTalker & 16.8\% & \underline{23.6\%} & \underline{17.8\%} \\
    DiffTalk & 12.9\% & 8.1\% & 16.5\% \\
    DreamTalk & 15.2\% & 10.6\% & 8.5\% \\
    \hline
    Ours & {\bf 26.6\%} & {\bf 53.8\%} & {\bf 51.6\%} \\
    \hline
    \end{tabular}
\end{table}

We conduct user studies with 20 participants to evaluate the performance of all methods. We generate 30 test videos covering different genders, ages, styles, and expressions. For each method, participants are required to choose the best one based on three metrics: 1) lip synchronization, 2) head motion diversity, and 3) overall image quality. As demonstrated in~\cref{tab:user_studies}, our work outperforms previous methods across all aspects, particularly in motion diversity and image quality. We attribute this to the decoupling of motion and appearance, as well as adopting diffusion models to generate higher-quality frames.
\section{Conclusion}

Talking head generation is an important research topic that still faces great challenges. Considering the issues of previous works, such as reliance on generative adversarial networks (GANs), regression models, and partial diffusion models, and neglecting the disentangling of complex facial representation, we propose a novel facial decoupled diffusion model, called FD2Talk, to generate high-quality, natural, and diverse results. Our FD2Talk fully leverages the strong generative ability of diffusion models and decouples the high-dimensional facial information into motion and appearance. We firstly utilize Diffusion Transformers to predict the accurate 3DMM expression and head pose coefficients from the audio signal, which serves as the decoupled motion-only information. Then these motion coefficients are fused into the Diffusion UNet, along with the appearance texture extracted from the reference image, to guide the generation of final RGB frames. Extensive experiments demonstrate that our approach surpasses previous methods in generating more accurate lip movements and yielding higher-quality and more diverse results.

\bibliographystyle{ACM-Reference-Format}
\bibliography{abbrev}

\end{document}